\documentclass[letterpaper, 10 pt, conference,]{ieeeconf}

\IEEEoverridecommandlockouts                              
\usepackage{graphicx}
\usepackage{siunitx}
\usepackage[utf8]{inputenc}
\usepackage[T1]{fontenc}
\usepackage{amsfonts}
\newcommand{\fig}[1]{Fig.~\ref{#1}}

\usepackage{epsfig} 
\usepackage{amsmath} 
\usepackage{amssymb}  
\usepackage{subcaption}
\usepackage{caption}
\usepackage{color}
\usepackage{verbatim}
\usepackage{wasysym}
\usepackage[table,xcdraw]{xcolor}
\usepackage{graphicx}
\usepackage{gensymb}
\usepackage{xcolor}
\usepackage{latexsym}
\usepackage{multirow}
\usepackage{amsmath}
\usepackage{booktabs,caption}
\usepackage[noadjust]{cite}
\usepackage[flushleft]{threeparttable}
\usepackage[table,xcdraw]{xcolor}

\usepackage[normalem]{ulem} 
\usepackage{wasysym}

\newcommand{\RNum}[1]{\uppercase\expandafter{\romannumeral #1\relax}}
\makeatletter
\newlength\tmp@\newlength\t@mp
\newcommand{\comp}[3]
  {\mathop{ \settowidth\tmp@{$\displaystyle\mathop{#1}^{#3}_{#2}$}
  \hbox to \tmp@{\hss \settowidth\t@mp{$\displaystyle #1$}\setlength\t@mp{.45\t@mp}
  $\displaystyle\mathop{#1}^{\hspace\t@mp #3}_{\hspace{-\t@mp}#2}$
  \hss} }}

\makeatother
\title{\LARGE \bf
Cycloidal Quasi-Direct Drive Actuator Designs with Learning-based Torque Estimation for Legged Robotics}

\author{Alvin Zhu$^{1*}$, Yusuke Tanaka$^{2*}$, Fadi Rafeedi$^{2}$ and Dennis Hong$^{2}$
\thanks{$^{1}$A. Zhu is with the Department of Computer Science and Electrical Engineering, $^{2}$Y. Tanaka, F. Rafeedi, and D. Hong are with the Department of Mechanical and Aerospace Engineering, UCLA, Los Angeles, CA, USA. \{alvister88, yusuketanaka, frafeedi, dennishong\}@g.ucla.edu.
$^*$A. Zhu and Y. Tanaka assert joint first authorship.
}%
}

\begin{document}
\maketitle


\begin{abstract}
This paper presents a novel approach through the design and implementation of Cycloidal Quasi-Direct Drive actuators for legged robotics. The cycloidal gear mechanism, with its inherent high torque density and mechanical robustness, offers significant advantages over conventional designs. By integrating cycloidal gears into the Quasi-Direct Drive framework, we aim to enhance the performance of legged robots, particularly in tasks demanding high torque and dynamic loads, while still keeping them lightweight. Additionally, we develop a torque estimation framework for the actuator using an Actuator Network, which effectively reduces the sim-to-real gap introduced by the cycloidal drive's complex dynamics. This integration is crucial for capturing the complex dynamics of a cycloidal drive, which contributes to improved learning efficiency, agility, and adaptability for reinforcement learning.
\end{abstract}

\section{Introduction}
Agile and dynamic locomotion in legged robots has been a longstanding challenge in robotics, requiring actuators that can deliver both dynamic torque control and high responsiveness, such as Direct Drive (DD) \cite{stanford_doggo}, \cite{DD_Legged_Robots} and Quasi-Direct Drive (QDD) \cite{cheetah} systems. 
However, integrating the gearbox into a confined space with conventional planetary, spur gear, and belt drive mechanisms is challenging without sacrificing gear load capacity since they are less resilient to significant impulse loads, such as those experienced during a fall.

Cycloidal gear mechanisms offer rigidity and superior gear teeth stress distribution \cite{app10041266}, allowing the design to be compact but promising adequate mechanical strength and minimal backlash \cite{Light_weight_CYC}, \cite{Cycloid_vs_Harmonic}. Integrating a cycloidal gearbox into a QDD system, the Cycloidal Quasi-Direct Drive (C-QDD) can enhance performance, especially in applications that demand high torque and dynamic load handling. However, cycloidal gears introduce nonlinearities in the torque output of the C-QDD, which can impact applications like reinforcement learning (RL) due to unmodeled dynamics in the actuator model during simulation \cite{Soni_2023}. Thus, modeling and estimating the C-QDD torque output is essential to bridging the sim-to-real gap \cite{actuator_net}.

Data-driven approaches for torque and force estimation, such as the Actuator Net \cite{actuator_net}, \cite{foot_gripper}, offer advantages since it is challenging to measure and account for manufacturing tolerances, to which cycloidal gears are particularly sensitive \cite{CYC_Error_Manufacture}. The significance of learning-based modeling lies in its ability to generalize across different operating conditions while maintaining high accuracy using only actuation history data \cite{YE2024115095}.

\begin{figure}
    \centering
    \includegraphics[width=0.85\linewidth]{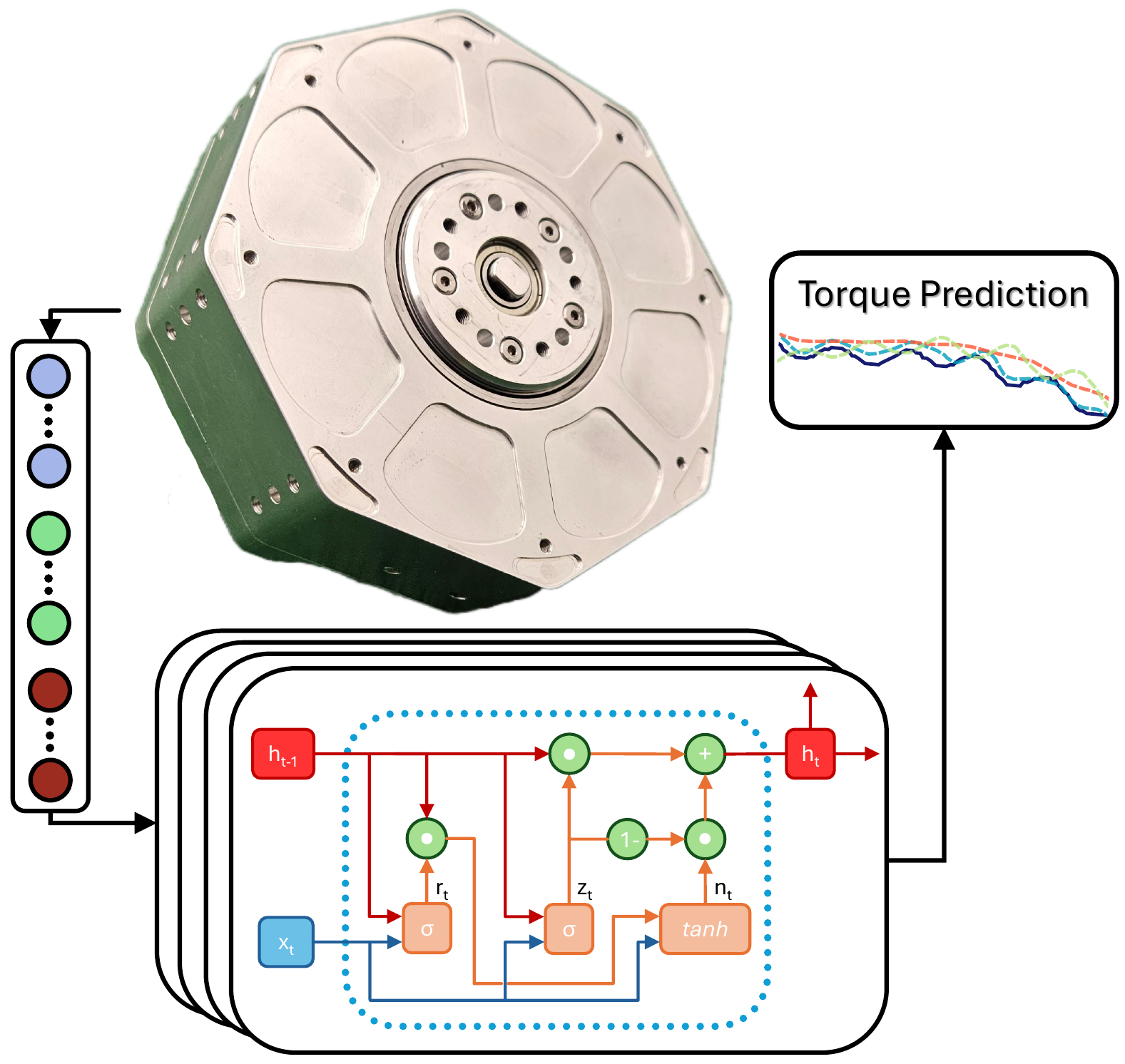}
    \captionsetup{width=0.85\linewidth}
    \caption{Proposed cycloidal gear quasi-direct drive actuator and the torque estimation framework. \label{fig:figure1}}
\end{figure}

This paper presents a QDD actuator with a 10:1 cycloidal gearbox for legged robots, shown in Fig. \ref{fig:figure1}. We also propose a torque estimation framework using gated recurrent units (GRU) \cite{chung2014empiricalevaluationgatedrecurrent} to model nonlinearities induced by the cycloidal gear, such as torque ripple. The C-QDD actuator is benchmarked and the performance of the torque estimation framework is evaluated experimentally.

The contributions of this paper are:

\begin{enumerate}
    \item \textbf{Design of a Cycloidal Quasi-Direct Drive Actuator}: We introduce a cycloidal quasi-direct drive actuator design tailored for dynamic locomotion and demanding climbing \cite{scaler-b}. 
    \item \textbf{Torque estimation framework}: We develop a GRU-based torque estimation framework that can predict non-linear, high-frequency features such as torque ripple to reduce the sim-to-real gap effectively.
    \item \textbf{Hardware verifications}: We evaluate the performance of the C-QDD and torque estimation framework through hardware experiments.  
\end{enumerate}

\begin{figure*}[htbp]
    \centering
    \begin{subfigure}[b]{0.74\linewidth}
        \centering
        \includegraphics[width=\linewidth, trim={0 0cm 0 0cm}, clip]{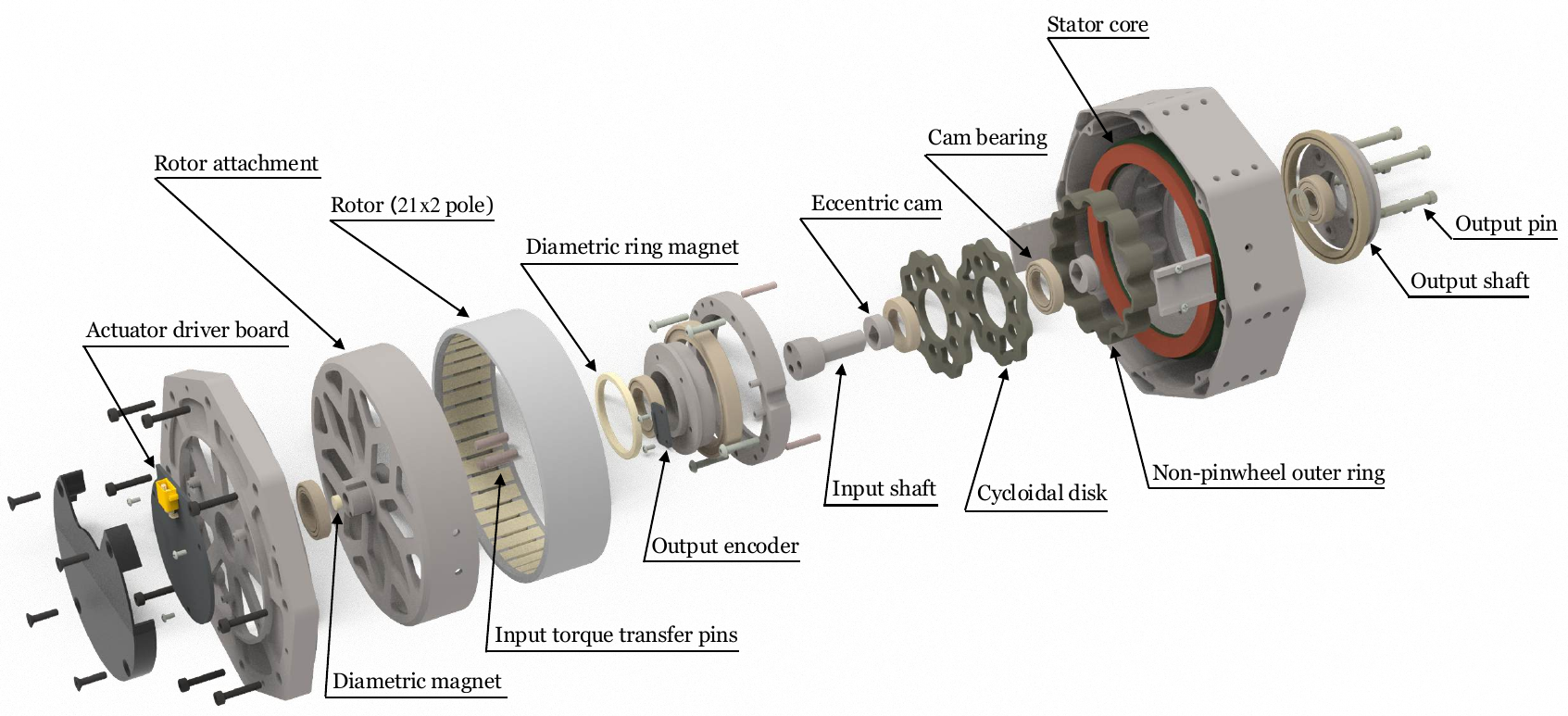}
        \caption{An exploded view of the C-QDD design.}
        \label{fig:exploded_view}
    \end{subfigure}
    \hfill
    \begin{subfigure}[b]{0.25\linewidth}
        \centering
        \includegraphics[width=\linewidth]{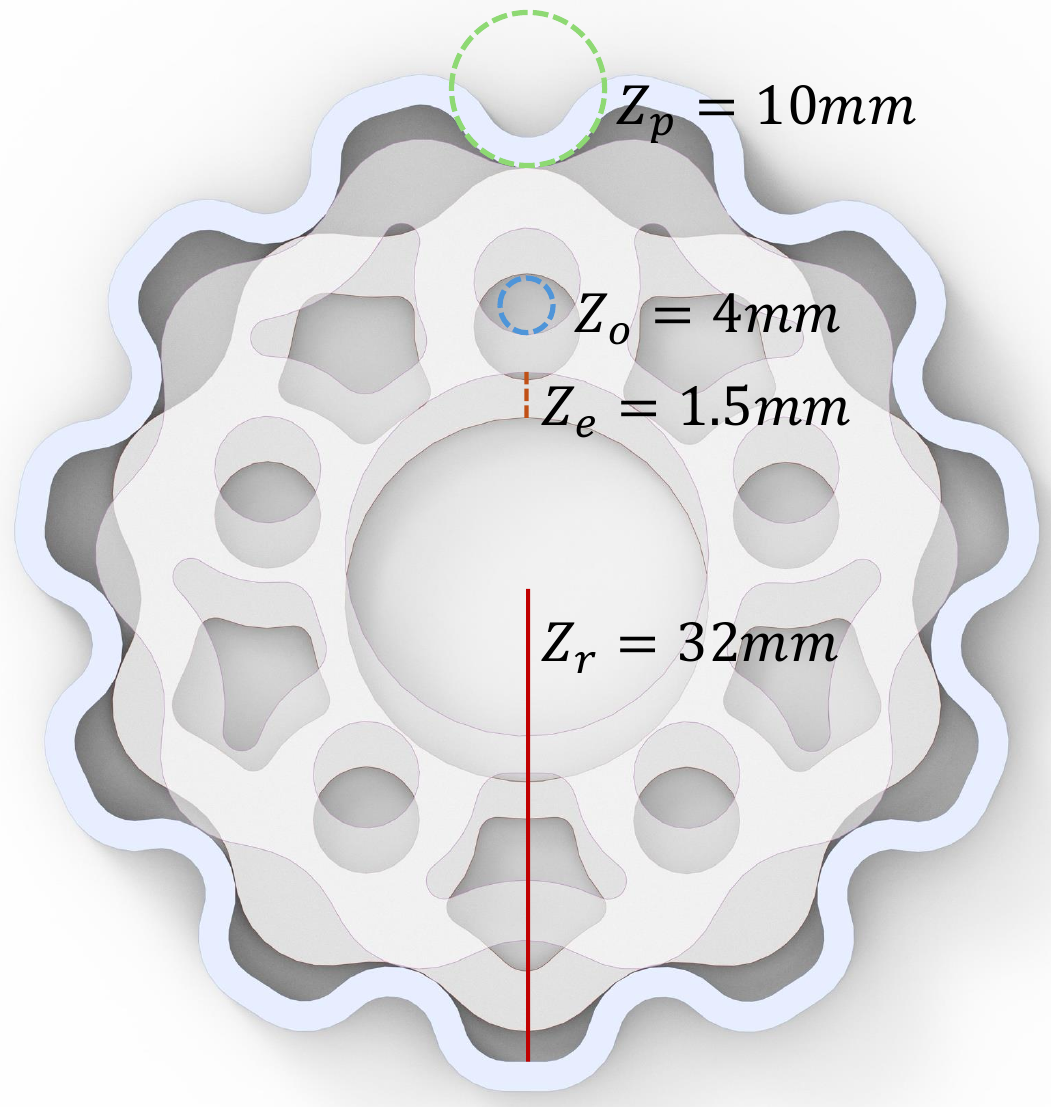}
        \caption{Cycloidal disk and ring profile with $Z_r$ as pitch radius, $Z_e$ as eccentricity, $Z_o$ as output pin diameter, and $Z_p$ as outer pin diameter.}
        \label{fig:cycloid_profile}
    \end{subfigure}
    \caption{C-QDD actuator design and cycloidal profiles.}
    \label{fig:cqdd_design}
\end{figure*}

\begin{figure}
    \centering
    \includegraphics[width=0.99\linewidth, trim={0cm 0cm 0cm 0cm}, clip]{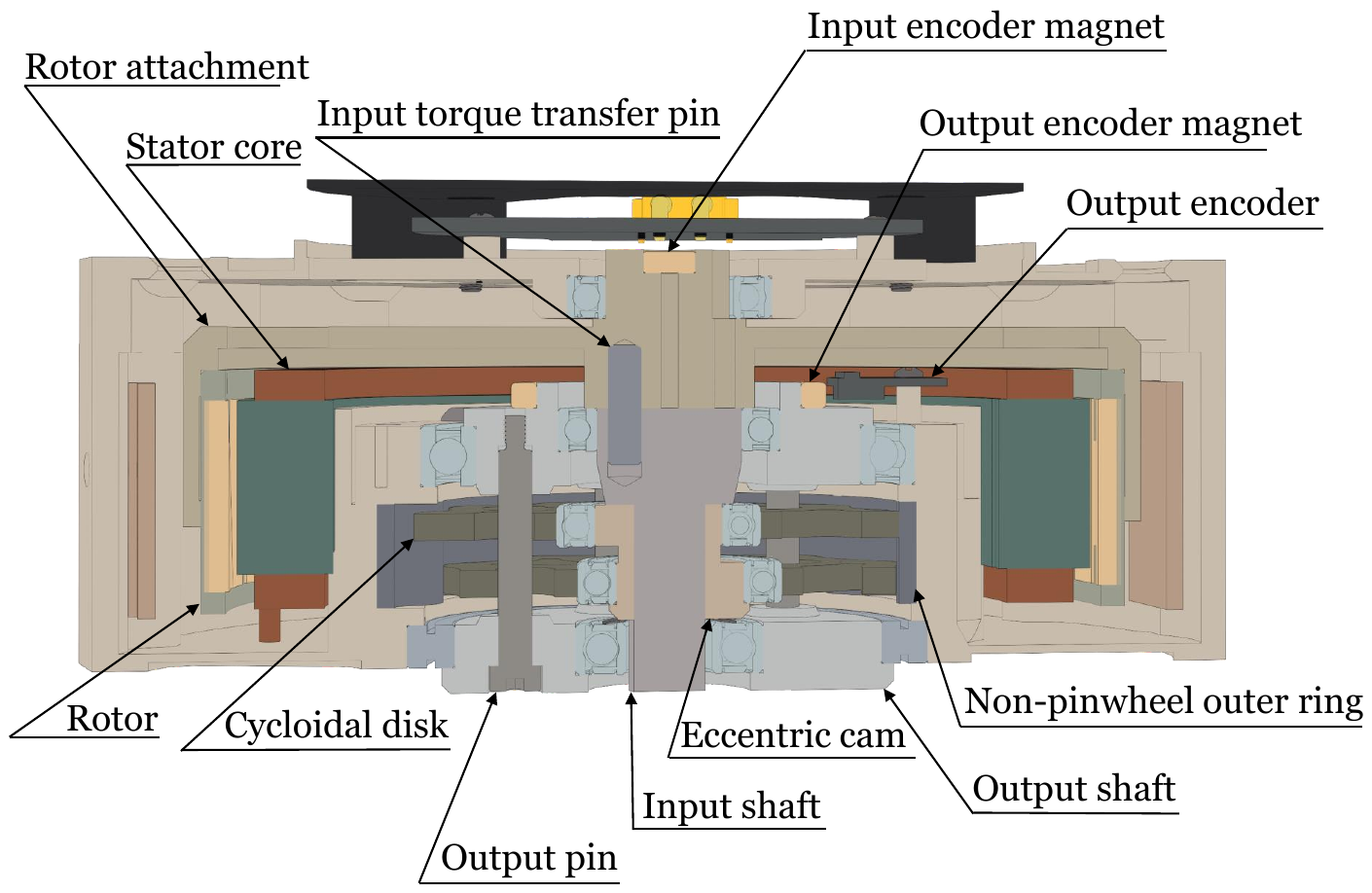}
    \caption{A cross-section view of the C-QDD design.}
    \label{fig:cross_section}
\end{figure}

\section{Related Works}

\subsection{Actuation Mechanism in Legged Robotics}
Traditional robotic actuators are designed to be rigid and for position control \cite{yusuke_scaler_2022}, \cite{GOAT}. However, rigid and non-backdrivable actuators are not ideal for dynamic and agile motion \cite{bruce}, \cite{back_drive_7dof_arm}, as compliance and torque control are essential in such scenarios \cite{cheetah}. 
Compliant actuators, such as series elastic \cite{Series_elastic_actuators} and variable stiffness actuators \cite{VIA_review}, address these challenges with a compliant spring element between the motor and the output. However, they require extensive modeling, optimization, and complex control algorithms.
Proprioceptive actuators, such as DD \cite{dd_leg} and QDD \cite{prismatic_qdd} actuators have been foundational in achieving high torque control fidelity and high responsiveness in legged robots \cite{biped_wheel}. DD actuators are praised for their simplicity and torque transparency, which minimize mechanical impedance \cite{cheetah} and lower reflected inertia. 
In contrast, QDD actuators integrate gear reductions,
while balancing torque amplification and maintaining the benefit of DD \cite{qdd_high_torque}. Recent advancements in QDD technology have focused on optimizing gear ratios and reducing backlash, further enhancing the actuator's performance in high-impact environments \cite{Pulse_actuator}.

\subsection{Cycloidal Gear Mechanisms}

Cycloidal gears have gained popularity as an alternative to planetary gears in legged robots due to their capacity to withstand large dynamic loads and frequent impacts \cite{Light_weight_CYC}, \cite{Cyc_Review}. Cycloidal gearboxes are compact and typically exhibit less backlash compared to planetary gears \cite{cyc_backlash}. However, despite these advantages, cycloidal gears also present several drawbacks stemming from the unique dynamics of the system \cite{machine_torque_ripple}. These include torque ripple, backlash, and unintended vibrations caused by the eccentricity of the rotors.

\subsection{Actuator Modeling}
 Actuator modeling and control are essential since the actuator dynamics
 inevitably contain non-linearity and dynamic variations \cite{system_id}. Actuator system identification is becoming increasingly important for addressing the sim-to-real gap in RL \cite{sim-to-real}, \cite{sim2real}.
 Traditional methods often struggle with these challenges, while data-driven approaches, such as actuator networks \cite{actuator_net}, have proven effective in improving the motor sim-to-real gap and torque prediction accuracy \cite{ANN_actuator}. 

Our C-QDD aims to benefit from the cycloid gear and quasi-direct drive mechanisms while compensating for the cycloid's complex dynamics by using a data-driven approach.

\begin{figure*}
    \centering
    \includegraphics[width=0.8\linewidth]{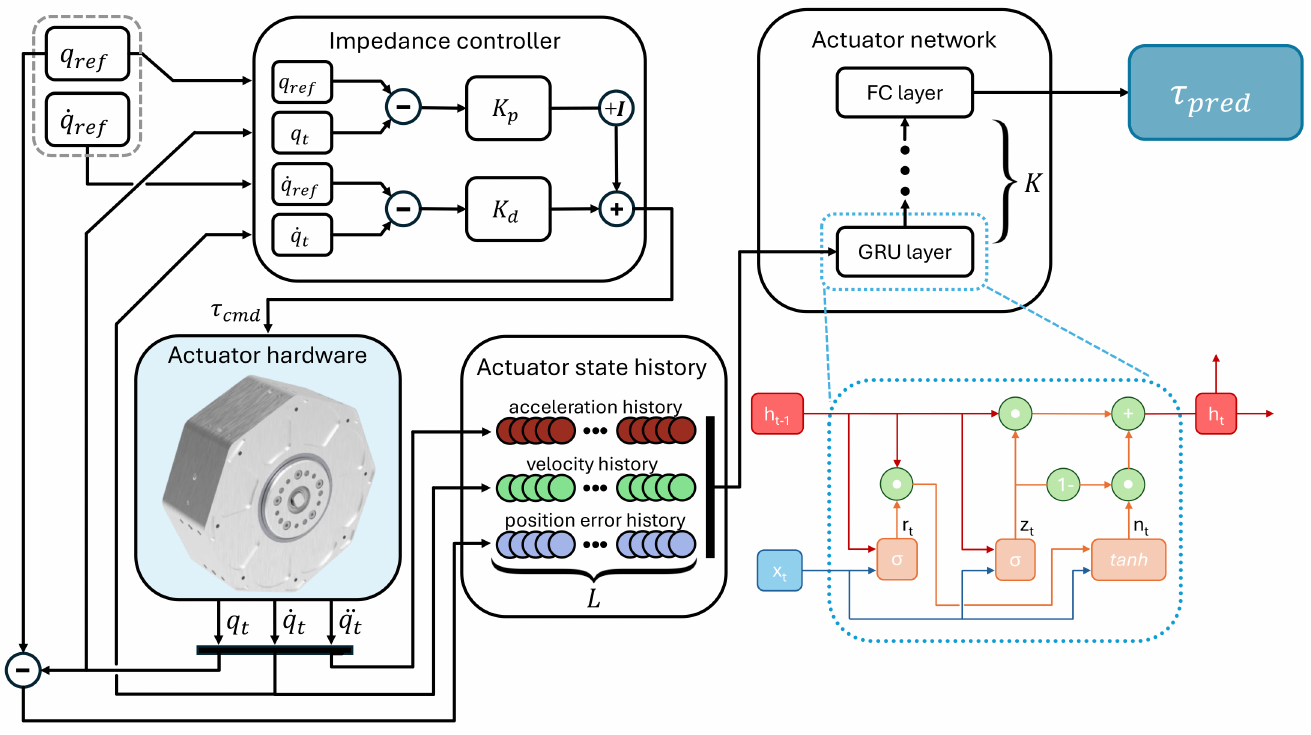}
    \caption{Overview of the torque estimation framework architecture, leveraging temporal \\ actuator state information to generate precise torque predictions.}
    \label{fig:network_pipeline}
\end{figure*}

\section{Cycloidal Gear Quasi-Direct Drive Mechanisms}

The use of cycloidal gears in low gear ratio and highly backdrivable actuators introduce unique challenges to the design of cycloidal mechanisms. In this section, we discuss the design principles behind C-QDD actuators displayed in Fig. \ref{fig:cqdd_design} and Fig. \ref{fig:cross_section}, focusing on how these parameters are optimized to achieve the desired performance characteristics.

\subsection{Sizing of Cycloidal and Planetary Gears}
Here, we compare the load capacity of our cycloidal gears to traditional planetary gears, which are common in proprioceptive actuators \cite{Light_weight_CYC}. The gearbox is integrated into the stator’s interior to maintain a compact design, limiting the maximum allowable pitch diameter for the cycloidal gear. A compound planetary gear can achieve a gear ratio of $10$ while remaining compact enough to fit within the stator. One such configuration includes a sun gear with $12$ teeth, planet gears with $36$ and $22$ teeth, and a ring gear with $70$ teeth.

For the given peak loads, the torque capacity of our cycloidal gear is limited by the output pins. In contrast, tooth bending and surface strength constrain the compound planetary gear's load capacity. To support the peak torque of $120$ Nm from the C-QDD, a planetary gear system would require six planets. This, in addition to not fitting within design constraints, will experience surface failure at the actuator's rated continuous torque levels of $40$ Nm for a module $1$ gear with a thickness of $12$ mm.

The cycloidal gear’s output pins are precision shoulder screws made of 18-8 stainless steel, positioned $19.5$ mm from the center of rotation. This configuration allows for a maximum torque of $59.13$ Nm per pin, providing a total of $295.65$ Nm of torque.

\subsection{Cycloid profiles and eccentricity}

The cycloid profile is controlled by the pitch diameter, $2Z_{r}$, eccentricity, $Z_e$, and outer pin diameter, $Z_{p}$, in Fig. \ref{fig:cqdd_design}b. 


The eccentricity of the cycloid gear induces an imbalance in gear inertia, which leads to undesired vibrations. To mitigate this effect, a counterbalancing disk is incorporated into the gearbox. The cycloid gear transmission ratio determines the number of counter disks required, $Z_R$, defined as:

\begin{equation}
Z_R = \frac{-Z_{np}}{Z_{np} - Z_{nt}} = -Z_{nt}
\end{equation}

where $Z_{np}$ represents the number of outer pins, and $Z_{nt}$ denotes the number of teeth on the cycloid gear. The number of disks, $N_Z$ required is denoted as:
\begin{align}
    Z_R = 2k \quad \text{(even number)}, \quad N_{Z} = 2 \\
    Z_R = 2k + 1 \quad \text{(odd number)}, \quad N_{Z} = 3
\end{align}

Here, $k \in \mathbb{Z}, \quad k\geq2$, and \(N_{\text{disks}}\) is the number of counterbalancing disks.

Two disks are sufficient to balance the inertia statically. However, to ensure dynamic stability, three disks are necessary to compensate for dynamic imbalances that may arise due to material deformation and manufacturing inaccuracy \cite{CYC_Error_Manufacture}. 
In the C-QDD design, two counter disks are used because the transmission ratio $Z_R = 10$ requires an even number of disks. However, this also significantly increases reflected inertia, which impacts the actuator's backdrivability.
The dynamic imbalance of the cycloid disk contributes to oscillations and torque ripples, which are compensated for through the torque estimation framework in Section \ref{sec:actuator_net}.

\subsection{Non-Pinwheel Cycloid Ring}
In the C-QDD, a non-pinwheel outer cycloid ring is employed, instead of the conventional pinwheel design \cite{pin-wheel_cyc}, \cite{non_pin-wheel_cyc}. Both the cycloid disks and the outer rings are manufactured from $4140$ alloy steel, known for their superior rigidity, wear resistance, and dimensional stability compared to the $7075$ aluminum housing.
Pinwheel designs traditionally utilize dowel pins embedded in the housing, which can present manufacturing challenges. Drilling pinholes directly into the housing necessitates tight tolerance control. C-QDD resolves these challenges by utilizing independent solid rings, which may accommodate better manufacturing tolerances among the cycloid disks, eccentric cams, and other components. The cycloid disks and rings are machined from tight tolerance ground flat stock using a wire EDM at $\pm 2 \, \mu\text{m}$ tolerance.

\section{C-QDD Dynamics and Torque Estimation}
To model the dynamics of the C-QDD, we employ a deep learning-based torque estimation framework, mitigating challenges such as force sensorless torque control, cycloid nonlinear ripple, and high-frequency oscillations. The torque estimation framework, depicted in \fig{fig:network_pipeline}, processes the actuator's state feedback to produce torque predictions.

\subsection{Torque Estimation Framework}

The proposed torque estimation framework utilizes a GRU-based architecture to model the actuator's dynamic behavior through the joint states. 
Over the time history horizon, $L$, two different sets of input vectors are defined as:

\begin{itemize}
    \item \textbf{PV-GRU}: $\mathbf{X}_{v_t} \in \mathbb{R}^{2 \times L}$, where $\mathbf{X}_{v_t} = [\mathbf{q_e}, \dot{\mathbf{q}}]^\top$
    \item \textbf{PVA-GRU}: $\mathbf{X}_{at} \in \mathbb{R}^{3 \times L}$, where $\mathbf{X}_{a_t} = [\mathbf{q_e}, \dot{\mathbf{q}}, \ddot{\mathbf{q}}]^\top$
\end{itemize}
The joint error $\mathbf{q}_e$ is calculated as $\mathbf{q}_e = \mathbf{q}_\text{ref} - \mathbf{q}$, where $\mathbf{q}_\text{ref}$ is the reference trajectory given and $\mathbf{q}$ is the current joint position. The $\dot{\mathbf{q}} \in \mathbb{R}^{L}$ represents the joint velocity over a history of $L$ timesteps.
The normalization of the inputs ensures consistent model performance and prevents scale imbalances that could hinder training.

The decision to include acceleration, $\ddot{\mathbf{q}} \in \mathbb{R}^{L}$, in the PVA-GRU variant arises to capture high-frequency, nonlinear torque dynamics more effectively—particularly phenomena such as torque ripple and oscillatory behavior. Such high-frequency dynamics require higher-order state information to enhance predictive accuracy, especially in regimes where position and velocity alone may be insufficient, as indicated by prior work \cite{actuator_net}. This will be further demonstrated through empirical validation in Section \ref{results}.

The GRU architecture was selected because of its capability to capture temporal dependencies efficiently, even across sequences of varying lengths. By maintaining an internal memory, the GRU processes historical state information \(x_t\) along with the hidden state \(h_{t-1}\) to iteratively update its predictions, thus enabling the network to learn long-term dependencies that are critical for accurately predicting high-frequency torque ripple.

\subsection{Model Parameters and Training}

The network is trained to minimize the error between the predicted torque, \(\tau_{pred}\), and the actual measured torque. Supervised learning is employed using historical data collected under diverse operating conditions of the C-QDD actuator, capturing a wide range of dynamic behaviors. 
The training pipeline uses the Adam optimizer with an initial learning rate (LR) of $1.0e^{-4}$, coupled with a One Cycle Learning Rate scheduler for efficient convergence \cite{smith2018superconvergencefasttrainingneural}, \cite{8624183}. This approach dynamically adjusts the LR to prevent overfitting while fine-tuning model performance during the final training stages. 


For our C-QDD, a $4$-layer stacked GRU with history size $30$ and batch size $64$ are selected.
The history size was determined by considering the dataset's sampling rate---200Hz---and the observed torque ripple frequency range. This ensures that the network receives enough historical data to capture the actuator’s intricate dynamics. Similarly, the GRU layers were tuned to introduce sufficient complexity in the network to model higher-order features while maintaining computational efficiency, crucial for real-time applications.

\section{Hardware Verification and Analysis\label{results}}
\begin{figure}
    \centering
    \includegraphics[width=0.8\linewidth, trim={0 0 0cm 5cm}, clip]{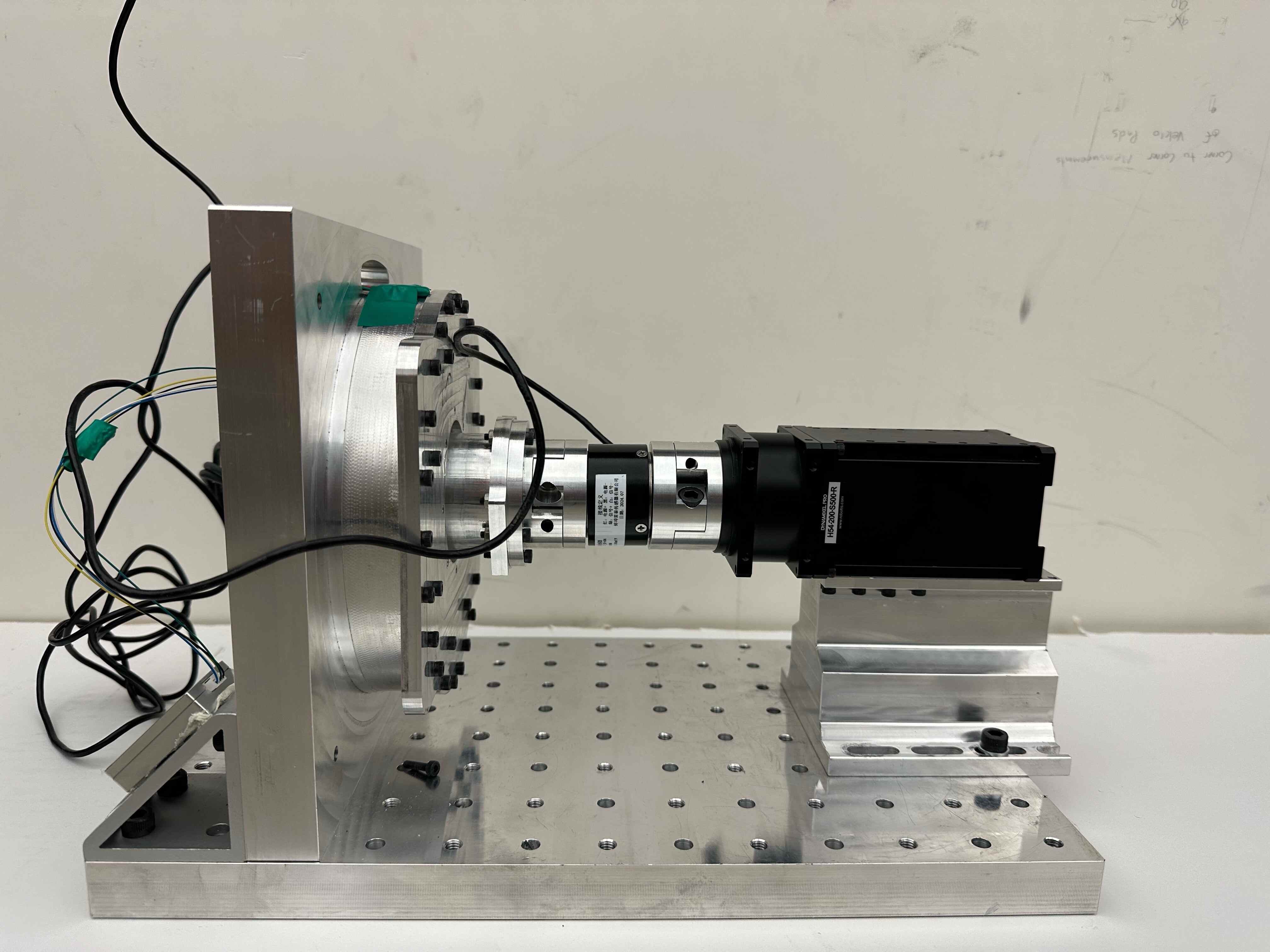}
    \caption{C-QDD actuator performance benchmark testbed.}
    \label{fig:testbed}
\end{figure}

In this section, we benchmark the C-QDD actuator's attributes and performance based on metrics critical to legged robots. These parameters include its torque and position control bandwidth, as well as its backdriving torque to highlight its ability to absorb impacts. The test set-up for our experiments is shown in \fig{fig:testbed}. In this test setup, the C-QDD's chassis is fixed, and the output is directly connected to a torque sensor, which is then connected to a Dynamixel Pro (H54-200-S500-R) servo motor. For tests requiring a high-impedance output setup, the servomotor is swapped with a rigid connection. 

\begin{figure}
    \centering
    \includegraphics[width=0.8\linewidth, trim={0cm 0cm 0cm 0cm}, clip]{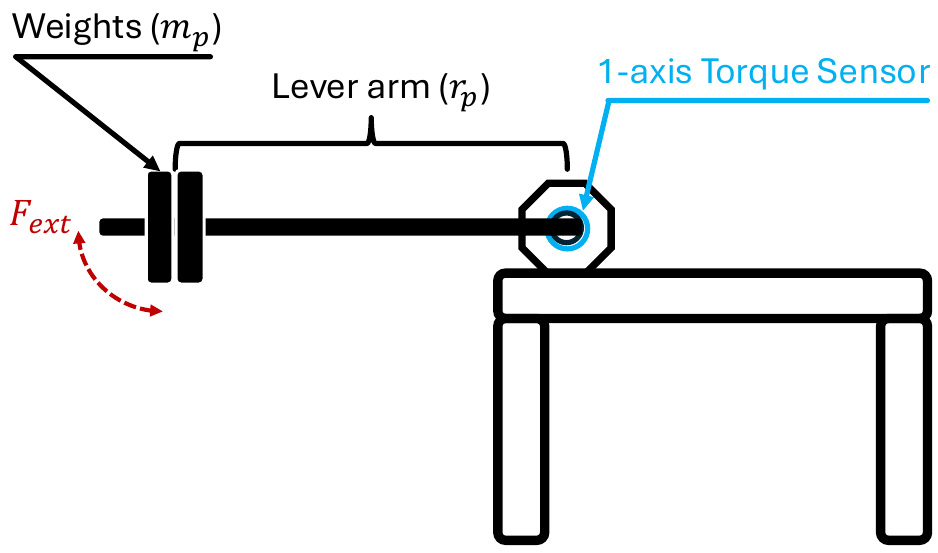}
    \captionsetup{width=0.8\linewidth} 
    \caption{Torque Estimation Framework data collection and performance testing setup.}
    \label{fig:net_test_setup}
\end{figure}

\subsection{C-QDD Mechanical and Electrical Performance}

C-QDD's continuous and peak torque values were measured using a high-impedance set-up and evaluated at $37.5$ Nm and $89.9$ Nm, respectively. The peak torque is lower than the expected peak torque of $120$ Nm based on the BLDC due to current limits on the C-QDD's CubeMars Driver Board-V2.1. Table. \ref{tb:T_Eff_comparison} compares C-QDD to some actuators intended for legged robots with similar features. A low amplitude torque tracking bandwidth was also experimentally obtained and evaluated at $34.3$ Hz. The torque control bandwidth was measured using the high-impedance setup by applying a $5$ Nm torque chirp signal to the actuator. The actuator output power was measured with a low-impedance dynamic torque transducer and encoder values. The peak electrical to mechanical output efficiency reached $82.3$\%.

 

\begin{table}[]
\renewcommand{\arraystretch}{1.2} 
\caption{C-QDD performance metrics}
\centering
\begin{tabular}{l|l}
\hline
 & \textbf{C-QDD} \\ \hline
\rowcolor[HTML]{EFEFEF} 
{Gear Type} & Cycloidal \\ 
{Gear Ratio} & 10 \\ 
\rowcolor[HTML]{EFEFEF} 
{Continuous Torque (Nm)} & 37.5 \\ 
{Peak Torque (Nm)} & 89.9 (120) \\ 
\rowcolor[HTML]{EFEFEF} 
{No Load Speed [24V] (rpm)} & 128.6 \\ 
{Rated Power (W)} & 960 \\ 
\rowcolor[HTML]{EFEFEF} 
{Output position Resolution (\degree)} & 0.00219 \\ 
{Efficiency (\%)} & 82.3 \\ 
\rowcolor[HTML]{EFEFEF} 
{Torque Density (Nm/kg)} & 64.21 (85.71) \\ 
{Actuator Inertia (kg$\cdot$m$^2$)} & $5.01 \cdot 10^{-4}$ \\ 
\rowcolor[HTML]{EFEFEF} 
{Weight (kg)} & 1.40 \\ 
{Backlash (arcmin)} & 7.03 $\pm$ 1.3 \\ 
\rowcolor[HTML]{EFEFEF} 
{Static Back-drive Torque (Nm)} & 1.99 \\ 
{Dynamic Back-drive Torque (Nm)} & 1.36 \\ 
\rowcolor[HTML]{EFEFEF} 
{Average Torque Ripple (Nm)} & $\pm$ 1.5 \\ 
{Torque Control Bandwidth [5 Nm] (Hz)} &  34.3 \\ 
\rowcolor[HTML]{EFEFEF} 
{Position Control Bandwidth [45$^\circ$] (Hz)} & 3.3 \\ 
{Position Control Bandwidth [5$^\circ$] (Hz)} & 22.13 \\ 
\end{tabular}
\end{table}

\begin{table}[]
\renewcommand{\arraystretch}{1.2} 
\centering
\caption{Efficiency and Torque Density Comparison}
\scriptsize 
\setlength{\tabcolsep}{2.5pt} 
\begin{tabular}{l|llll}
\hline
Metric                     & C-QDD          & BEAR \cite{BEAR_Tym} & PULSE115-60 \cite{Pulse_actuator} & Lee et al. \cite{Light_weight_CYC}
\\ 
\rowcolor[HTML]{EFEFEF} 
\hline
Peak Torque (Nm)           & 89.9          & 32              & 62.5               & 155.32       
\\ \hline
Mass (kg)                  & 1.40          & 0.670           & 1.25               & 1.85         
\\ \hline
\rowcolor[HTML]{EFEFEF} 
Gearing Ratio              & 10:1          & 10:1            & 5:1                & 11:1       
\\ \hline
Torque Density (Nm/kg)     & 64.2          & 47.7            & 50                 & 83.7        
\\ \hline
\rowcolor[HTML]{EFEFEF} 
Efficiency (\%)            & 82.3          & -               & -                  & 90         
\\ \hline
\end{tabular}
\label{tb:T_Eff_comparison}
\end{table}

\subsection{C-QDD Backlash and Backdrivability Performance}

\subsubsection{Backlash}
The backlash of the C-QDD was measured based on the method discussed in \cite{gebler1998identification}. For this test, the C-QDD chassis was fixed to the bed with no load attached to its output. The actuator is first preloaded in the opposite direction, and then the BLDC motor is gradually rotated until an output shaft motion is observed. 
A dial indicator with an accuracy of $0.0127$ mm was used to identify when the output shaft moved. 
To account for inaccuracies in the manufacturing and assembly processes, this test was performed at six different locations of the output shaft, preloaded in both directions. The average backlash was calculated at $7.0$ arcminutes with an uncertainty of $\pm 1.3$ arcminutes. The backlash is comparable to other actuators designed for similar purposes, such as the one in \cite{Pulse_actuator}.


\subsubsection{Static and Dynamic Backdriving Torque\label{sec:static_backdrive}}
The torque needed to back-drive the C-QDD from the output was measured through the test setup shown in \fig{fig:testbed}. The servo motor was used to back-drive the output of the C-QDD with a torque sensor rigidly fixed concentrically between the outputs of the actuator and servo. The static and dynamic backdriving torques—required to initiate and maintain motion, respectively—were both evaluated. The static and dynamic backdriving torques were observed at $1.99$ Nm and $1.36$ Nm, respectively. When compared to other QDD actuators, such as \cite{Pulse_actuator} and \cite{Quasi-Hip_Exo}, which have backdriving torques of $0.37$ Nm and $0.97$ Nm, the backdriving force is higher, but comparable considering the C-QDD incorporates a higher gear-ratio and has a considerately higher torque capability.  


\begin{figure*}
    \centering
    \includegraphics[width=0.92\linewidth]{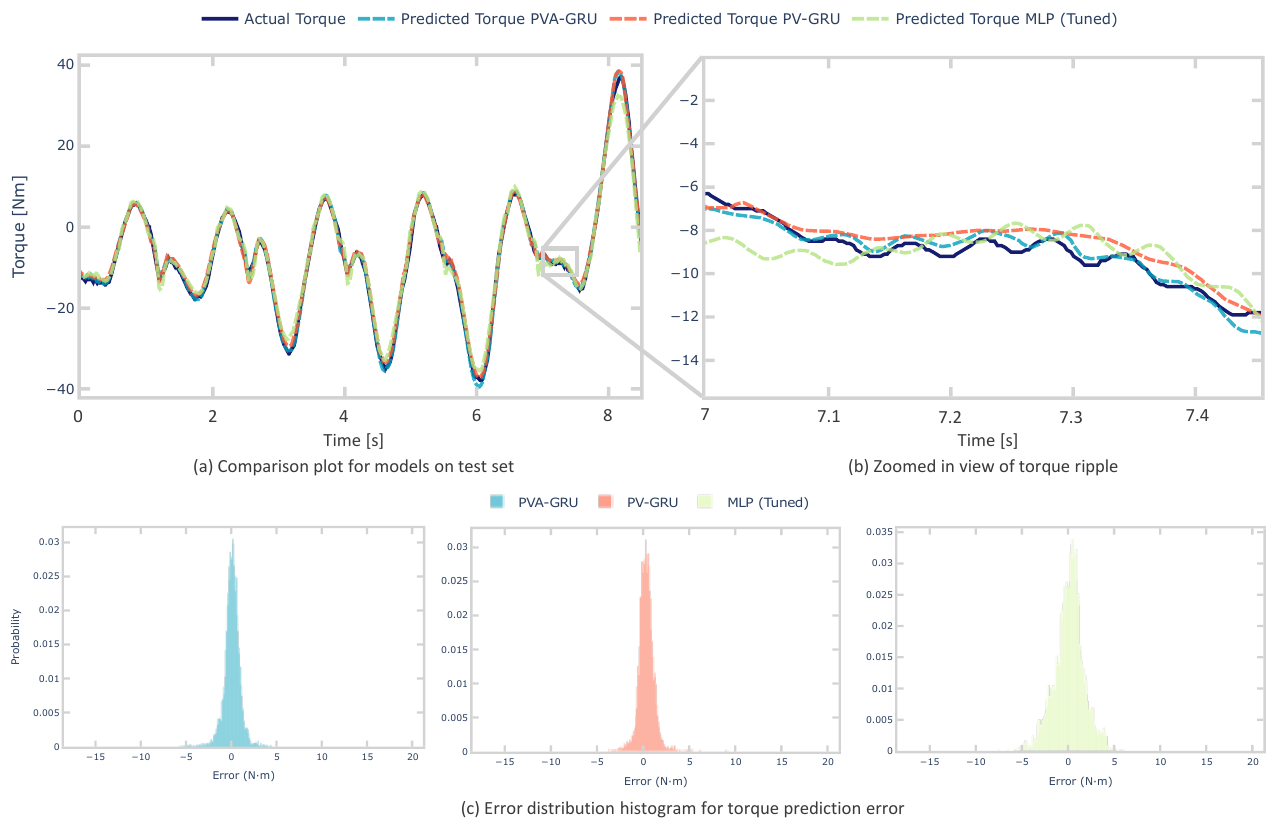}
    \caption{The graphs show the estimated torque for all the models compared to the ground truth torque during live inference, \\ along with their variances.}
    \label{fig:compare_graph}
\end{figure*}


\subsection{Torque Estimation and Actuator Network Performance}\label{sec:actuator_net}

To evaluate the Actuator Network's performance for torque estimation, $\mathcal{X} = \{q, \dot{q}, \ddot{q},\tau_q\}$ were collected on a single inverted pendulum setup using our C-QDD shown in Fig. \ref{fig:net_test_setup}. The training data was generated using sinusoidal joint angle profiles with varying frequencies $f_q\in[0.3, 1.5]$ Hz, amplitude $\mathcal{A}_q\in[0.5, 1.5]$ rad, and initial position $q_{t=0} \in [0, 1]$ rad. The pendulum mass, $m_p\in \{1.14, 2.28\}$ kg was attached at varying locations on the pendulum bar, $r_p\in[0.25, 0.6]$ m, which represents various loads and inertia of the pendulum arm. The pendulum is allowed to make contact with a compliant wall, simulating the hybrid dynamics characteristics of legged robotic motion.

\begin{table}[t!]
\centering
\caption{Comparison of Actuator Networks}

\setlength{\tabcolsep}{2pt} 
\renewcommand{\arraystretch}{1.2} 
\begin{tabular}{l|lllll}
\hline
Metric                     & \textbf{PVA-GRU}          & \textbf{PV-GRU}           & \textbf{MLP (Tuned)}      & \textbf{MLP (Baseline)}
\\ 
\rowcolor[HTML]{EFEFEF} 
\hline
RMSE (Nm)                  & 0.97          & 1.00           & 1.71          & 2.46              
\\ \hline
Variance (Nm)              & 0.92          & 0.85           & 2.90           & 3.62        
\\ \hline 
\rowcolor[HTML]{EFEFEF} 
Cycle Time (\(\mu\)s) & 1.49  & 3.10      & 0.72         & 0.22
\\ \hline
Layers                      & 4             & 4                & 3               & 3                   
\\ \hline
\rowcolor[HTML]{EFEFEF} 
History Size                & 30            & 30               & 24              & 3                
\\ \hline
\end{tabular}
\label{tb:network_comparison}

\vspace{0.3em} 
\footnotesize{\parbox{\linewidth}{\raggedright * Cycle time refers to the execution time on the fastest device; device is CPU for all models; all models use a hidden size of 32.}}

\end{table}

\subsubsection{Actuator Net Performance Comparisons}

\fig{fig:compare_graph} graphs our torque estimation models, PVA-GRU and PV-GRU, against the ground truth. Root Mean Square Error (RMSE) and statistical torque estimation error distributions are shown in \fig{fig:compare_graph}c. Table. \ref{tb:network_comparison} compiles RMSE and variances for all models. 

As a baseline, a multilayer perceptron (MLP) used in \cite{actuator_net} was trained on our dataset, and a tuned version of the MLP was also trained for further comparison. The tuned MLP and baseline MLP had an RMSE of $1.71$ Nm and $2.46$ Nm, respectively, demonstrating comparable results to the original implementation in \cite{actuator_net}, which is applied to a serial elastic actuator instead of a cycloidal gear actuator and operates under smaller torque ranges. \fig{fig:compare_graph}a shows a visualization where all models agree with the ground truth for large-scale torque estimation. However, in \fig{fig:compare_graph}b, the tuned MLP could not accurately capture small torque variation features, such as sinusoidal torque ripple with $0.6$ Nm amplitude and $20$ Hz frequency, showing a phase shift behavior of almost $180\degree$.

On the other hand, the PVA-GRU model effectively leverags the higher-order acceleration data to predict small amplitude and high-frequency variations accurately. As shown in \fig{fig:compare_graph}b, the PVA-GRU estimated the torque ripple amplitude with an average of $0.23$ Nm error from the ground truth, with a frequency of $17.6$ Hz, compared to the ground truth of $17.2$ Hz, and a phase shift of less than $40\degree$. Furthermore, the model demonstrated a significant improvement of $43.5\%$ in RMSE and $68.2\%$ in variance compared to the tuned MLP. Accurately capturing such torque nonlinearities is crucial for applications like RL simulations, where minor discrepancies in torque estimation can impact performance \cite{Soni_2023}.

The PV-GRU torque estimations resemble a damped version of the ground truth output, with an average torque ripple prediction error of $0.71$ Nm as shown in \fig{fig:compare_graph}b. 
One notable drawback of the PVA-GRU is that it relies on the noisier acceleration measurements. Thus, the predicted mean over time becomes less smooth compared to PV-GRU as shown in both \fig{fig:compare_graph}a and the larger variance shown in  Table. \ref{tb:network_comparison}. Nonetheless, statistically, the PVA-GRU has shown the highest accuracy overall.  
Hence, this supports that the PVA-GRU approach is effective for modeling actuator torque with detailed nonlinear features.

\subsubsection{Computational Cost}

In terms of computational cost, the PVA-GRU and PV-GRU run at $1.5$ $\mu$s and $3.1$ $\mu$s, respectively, on CPU, while the MLPs run at $<1.0$ $\mu$s. On GPU, the PVA-GRU and PV-GRU run at $13.0$ $ \mu$s, and the MLPs run at $4.2$ $\mu$s, including CPU-GPU data transfer overhead. The models slow down on the GPU due to their simple network structure, where data transfer overhead becomes a bottleneck. These benchmarks were measured on an Intel Core i9-13900K CPU and an Nvidia RTX A2000 GPU on a test set of 16,000 data points.

The GRU-based architecture achieves real-time performance during live hardware verification, with average forward passes of 2 $\mu$s on CPU. Such processing speed is crucial for dynamic control tasks like legged locomotion, where fast response times ensure control loop frequencies above 1 kHz, necessary for torque control and quick terrain adaptation. In GPU-accelerated simulations, training avoids CPU-GPU transfer overhead, allowing models to run at speeds comparable to CPU inference. Accurate torque modeling, combined with GPU-accelerated parallel RL environments, allow for fast training with better sim-to-real transfer, improving zero-shot hardware deployment of learned policies.

By incorporating acceleration into the state-space representation, the PVA-GRU model demonstrated accuracy improvements over the baselines and the PV-GRU in terms of predictive capability, particularly in the presence of high-frequency nonlinearities, solidifying its utility in precision torque control for advanced robotic systems.

\section{Conclusion}

This paper introduces a novel Cycloidal Quasi-Direct Drive actuator and torque estimation framework designed to meet the demanding needs of agile-legged robotics and climbing. The C-QDD actuator offers high torque density, mechanical rigidity, and minimal backlash, making it exceptionally effective for high-load, dynamic conditions commonly encountered in legged robotics. Our GRU-based torque estimation framework demonstrates the ability to model the actuator’s complex torque characteristics across high and low-amplitude motions without relying on direct torque sensing, potentially bridging the gap between simulation and real-world performance.

The hardware benchmarks have validated the C-QDD’s superior torque density, minimal backlash, and robust handling of dynamic loads, positioning it as a powerful solution for dynamic robotic systems. Our future research will focus on integrating this actuator and torque estimation framework into force/torque-aware planning \cite{contact_rich}, control \cite{alex_admittance}, and RL applications. By leveraging the torque estimation framework's high degree of accuracy and efficient computation, we aim to integrate the framework with both training and hardware for better sim-to-real transfer, ultimately enhancing the peformance of legged robots in challenging scenarios.



\newpage

\bibliographystyle{IEEEtran}
\bibliography{main,scaler}

\end{document}